\documentclass[conference]{IEEEtran}

\usepackage{cite}
\usepackage{graphicx}
\usepackage{booktabs}
\usepackage{array}
\usepackage{url}
\usepackage{xcolor}
\usepackage{amsmath}
\usepackage{balance}

\title{Quantifying the Gap Between Laboratory Battery Test Patterns and Field Duty Profiles}

\author{
\IEEEauthorblockN{Chunyang Zhao and Chresten Tr{\ae}holt}
\IEEEauthorblockA{
Department of Wind and Energy Systems\\
Technical University of Denmark\\
Kongens Lyngby, Denmark\\
chuzh@dtu.dk, ctra@dtu.dk}
}

\begin{document}
\maketitle

\begin{abstract}
Laboratory battery tests provide the main empirical basis for battery performance and degradation studies, but their operating patterns do not directly represent field duty profiles. This paper quantifies the gap by comparing six accessible evidence sources covering controlled cycling, drive-cycle testing, dynamic cycling, NMC811 laboratory ageing, a real electric-vehicle charging trace, and fleet-scale electric-vehicle state-of-health (SOH) data. The analysis combines usage frequency, usage intensity, usage C-rate, and a duty-structure index (DSI) based on normalized current dispersion and ramping. The representative single-segment DSI ranges from 0.630 for the field source trace and 0.699 for NASA to 2.936 for Oxford and 2.855 for Imperial, while usage C-rate ranges from 0.14--0.40 for Imperial, NASA, Stanford, and Hyundai to 2.00 for Oxford. Long-term ageing also differs: the 80\% retention region occurs near 351 NASA cycles, 6292 Oxford checkpoints, and 1019 Stanford cycles. In chemistry-aligned NMC/NCM evidence, Imperial retains 0.813 under standard cycling and 0.865 under drive-cycle ageing, while the field source has median SOH 0.889 with visible dispersion. Field operation further shows a median use intensity of 137.2~km/day and 56.9\% of charges ending at or above 95\% SOC. These results show that battery performance metrics are conditional on the duty pattern that generated them; application-oriented studies should report explicit duty-profile descriptors together with chemistry, capacity, and ageing metrics.
\end{abstract}


\section{Introduction}

Battery cells and battery energy storage systems are now deployed across electric vehicles, stationary storage, microgrids, hybrid renewable systems, and grid-support applications~\cite{zhao2023bess}. In each of these settings, battery behaviour is shaped not only by chemistry and cell design, but also by the operating pattern imposed over time: current transients, voltage windows, state-of-charge (SOC) limits, temperature exposure, rest periods, and supervisory control all influence usable energy, efficiency, thermal stress, and ageing. The consequence is simple but important. Battery performance is inseparable from the duty profile~\cite{geslin2025dynamic}.

Most battery knowledge, however, is still generated in the laboratory. Although fundamental material and reaction studies continue to advance battery science, controlled cycling, constant-current constant-voltage charging, periodic reference tests, and temperature-regulated ageing campaigns remain indispensable because they provide repeatability, interpretability, and a defensible basis for cross-study comparison. Open datasets from NASA PCoE~\cite{nasa}, Oxford~\cite{oxforddataset,birkl}, the early-life benchmark introduced by Severson \emph{et al.}~\cite{severson}, and more recent dynamic-protocol studies, such as the Stanford work~\cite{stanford}, have greatly expanded the empirical basis for battery research. Yet the operating logic of these tests remains more structured than that of deployed batteries.

This mismatch creates a transfer problem. A battery result may be valid under a laboratory protocol and still be only partially informative for a real application. In practice, field duty profiles are stochastic, partially observed, and strongly shaped by battery-management systems, charger behaviour, route conditions, driver behaviour, and system-level constraints. Recent open field-EV work has made this distinction clearer by showing both the richness of real operating histories and the difficulty of translating laboratory-derived health indicators directly into real-world battery management~\cite{fieldsource}. The issue is not that laboratory testing is inadequate; rather, different test patterns support different kinds of claims.

This paper addresses this issue by comparing battery evidence at the test pattern level. We propose two linked research questions: which battery-performance statements are supported by controlled, application-inspired, dynamic, and field-derived evidence, and how far do those evidence types remain from the actual field duty structure? To answer them, we assemble a compact but reproducible comparison spanning controlled cycling, drive-cycle-inspired cell tests, dynamic cycling, chemistry-aligned evidence, a real EV charging trace, and field state-of-health statistics.

The key contribution of the paper is not a new dataset or a new lifetime model, but a structured way to interpret battery performance evidence across the laboratory-to-field gap. Specifically, the paper contributes: 1) a duty-pattern-based comparison framework that treats battery datasets as evidence classes rather than interchangeable benchmarks; 2) a compact quantitative comparison spanning transient response, operating variability, and long-term retention across accessible commercial lithium-ion battery testing and operation sources; and 3) a quantitative analysis approach to compare and analyze the testing pattern of the battery cell operations. After the introduction sections, the remainder of the paper is organized as follows: Section~II defines the evidence base and workflow, Section~III presents the comparison results, Section~IV discusses evidence limits, and Section~V concludes the paper.

\begin{table*}[tb]
\caption{Focused data description for the quantified sources used in the comparison.}
\label{tab:data_description}
\centering
\footnotesize
\setlength{\tabcolsep}{2pt}
\renewcommand{\arraystretch}{1.08}
\begin{tabular*}{\textwidth}{@{\extracolsep{\fill}}
>{\raggedright\arraybackslash}p{2.25cm}
>{\raggedright\arraybackslash}p{1.00cm}
>{\raggedright\arraybackslash}p{1.35cm}
>{\raggedright\arraybackslash}p{1.35cm}
>{\raggedright\arraybackslash}p{2cm}
>{\raggedright\arraybackslash}p{2cm}
>{\raggedright\arraybackslash}p{2cm}
>{\raggedright\arraybackslash}p{5.20cm}
@{}}
\toprule
Source
& System
& Chemistry
& Capacity
& $V$ mean [min,max] (V)
& $I$ mean [min,max]
& $T$ mean [min,max] ($^{\circ}$C)
& Test / operating case \\
\midrule

NASA B0005~\cite{nasa}
& Cell
& Li-ion
& 2.0 Ah
& \begin{tabular}[t]{@{}l@{}}3.56\\{[2.63, 4.20]}\end{tabular}
& \begin{tabular}[t]{@{}l@{}}-1.92 A\\{[-2.02, 0.00] A}\end{tabular}
& \begin{tabular}[t]{@{}l@{}}31.92\\{[23.78, 38.46]}\end{tabular}
& Controlled degradation baseline; constant-current ageing with capacity and impedance follow-up \\

Oxford NCA Cell1~\cite{oxforddataset,birkl}
& Cell
& NCA
& 0.74 Ah
& \begin{tabular}[t]{@{}l@{}}3.88\\{[3.60, 4.20]}\end{tabular}
& \begin{tabular}[t]{@{}l@{}}-0.12 A\\{[-5.00, 1.60] A}\end{tabular}
& \begin{tabular}[t]{@{}l@{}}40.38\\{[39.93, 41.09]}\end{tabular}
& Application-shaped bridge dataset; drive-cycle discharge in thermal-chamber testing \\

Imperial 21700~\cite{imperial}
& Cell
& NMC811 / C-SiOx
& 4.865 Ah
& \begin{tabular}[t]{@{}l@{}}3.66\\{[2.50, 4.17]}\end{tabular}
& \begin{tabular}[t]{@{}l@{}}-0.50 A\\{[-0.50, -0.50] A}\end{tabular}
& \begin{tabular}[t]{@{}l@{}}25.93\\{[24.35, 26.70]}\end{tabular}
& Chemistry-aligned ageing reference; standard-cycle vs drive-cycle ageing at $25\,^{\circ}\mathrm{C}$ \\

2025 EV fleet source~\cite{fieldsource}
& EV fleet
& NCM
& 155 Ah\textsuperscript{a}
& \begin{tabular}[t]{@{}l@{}}4.04\\{[3.75, 4.25]}\end{tabular}
& \begin{tabular}[t]{@{}l@{}}73.61 A\\{[0.00, 103.50] A}\end{tabular}
& \begin{tabular}[t]{@{}l@{}}31.25\textsuperscript{b}\\{[30.00, 32.00]\textsuperscript{b}}\end{tabular}
& Field health and usage diversity; in-service EV operation, charging windows, and SOH statistics \\

Stanford cell 003~\cite{stanford}
& Cell
& SiOx--Gr / NCA
& n.r.\textsuperscript{c}
& \begin{tabular}[t]{@{}l@{}}3.57\\{[2.98, 4.20]}\end{tabular}
& \begin{tabular}[t]{@{}l@{}}0.00\textsuperscript{d}\\{[-0.10, 0.50]\textsuperscript{d}}\end{tabular}
& \begin{tabular}[t]{@{}l@{}}35.00\textsuperscript{e}\\{[35.00, 35.00]\textsuperscript{e}}\end{tabular}
& Dynamic protocol comparison; long charge-rest-discharge sequence under dynamic cycling \\

Hyundai Ioniq 5~\cite{kitdataset}
& Vehicle session
& SK On NCM pouch
& 111--121 Ah\textsuperscript{f}
& \begin{tabular}[t]{@{}l@{}}233.29\\{[75.48, 285.19]}\end{tabular}
& \begin{tabular}[t]{@{}l@{}}30.69 A\\{[0.52, 48.08] A}\end{tabular}
& \begin{tabular}[t]{@{}l@{}}2.42\textsuperscript{g}\\{[-0.50, 5.40]\textsuperscript{g}}\end{tabular}
& Real duty-trace example; real EV AC charging session with SOC and phase-power records \\

\midrule
\multicolumn{8}{@{}p{\textwidth}@{}}{
\textsuperscript{a} Liu \emph{et al.} report a rated pack capacity of 155 Ah with 96 cells connected in series.
\textsuperscript{b} EV-fleet temperature is the charging-temperature health-indicator sequence from the source-data release.
\textsuperscript{c} Stanford reports commercial SiOx--graphite/NCA EV energy cells and normalized capacity/C-rate data; the exact cell model and absolute nominal capacity are not disclosed.
\textsuperscript{d} Stanford current is reported in normalized C-rate rather than amperes.
\textsuperscript{e} Stanford temperature is the chamber set point reported for the ageing campaign.
\textsuperscript{f} KIT leaves the selected Hyundai capacity field blank; Hyundai battery-cell information lists Ioniq~5 cells as SK~On pouch-type NCM, while public pack specifications imply about 111--121~Ah across the 72.6/77.4/84~kWh variants.
\textsuperscript{g} KIT temperature is ambient temperature during the selected Hyundai charging session.} \\
\bottomrule
\end{tabular*}
\end{table*}

\section{Study Design and Evidence Base}
Table~\ref{tab:data_description} summarizes the selected evidence base for comparing battery test patterns and field duty, and the representative segments of duty cycle in the normalized current profiles are presented in Fig~\ref{fig:patterns}. The table identifies the system level, chemistry, nominal capacity, signal range, and scientific role of each source. This distinction is necessary because the selected records span sub-Ah laboratory cells, commercial 21700 cells, large EV cells or packs, and fleet-level source data; their voltage and current magnitudes, therefore, cannot be pooled directly. The comparison is organized by the claim each source can support: controlled response and fade for NASA, application-shaped cell cycling for Oxford, chemistry-aligned NMC811 ageing for Imperial, fleet SOH and usage dispersion for the 2025 EV source, sequence-rich dynamic cycling for Stanford, and raw field charging structure for the Hyundai Ioniq~5 session.

The first layer of quantification follows the long-term usage-description method proposed in our prior work~\cite{zhao2023bess}. Conventional battery descriptors such as chemistry, nominal capacity, voltage, current, and temperature describe hardware properties or instantaneous operating states, but they do not fully describe how a battery is used over an application period. For this reason, usage frequency ($UF$), usage intensity ($UI$), and usage C-rate ($UC$) are used to describe how often the battery is active, how many cycles are accumulated during active use, and how large the charging-current level is during use.

\begin{equation}
\mathrm{Usage\ frequency} =
\frac{\mathrm{Active\ usage\ time\ length}}
{\mathrm{Application\ time\ length}}
\label{eq:uf}
\end{equation}

\begin{equation}
\mathrm{Usage\ intensity} =
\frac{\mathrm{Cycle\ count}}
{\mathrm{Active\ usage\ time\ length}}
\label{eq:ui}
\end{equation}

\begin{equation}
\mathrm{Usage\ C\mbox{-}rate} =
\frac{\int |\mathrm{C\mbox{-}rate}|\,dt}
{\mathrm{Active\ charging\ time\ length}}
\label{eq:uc}
\end{equation}

Here, active usage time is the accumulated charging and discharging time, application time is the total elapsed application period including standby, and active charging time is the charging portion used for the usage C-rate calculation. The cycle count can be obtained through rainflow counting or another consistent equivalent-cycle method. In this paper, SOC is not available for all sources, so cycle count is estimated consistently from current throughput when a direct cycle count is not available. For Stanford, the reported normalized C-rate is adopted directly. For the 2025 EV source-data trace, sample-coordinate normalization is used because timestamps are unavailable. For Hyundai, the value is treated as an AC-current proxy because the KIT charging trace does not provide direct battery DC current.

These three usage metrics provide the long-period interpretation, but they do not by themselves capture short-period current shape. Each selected record is therefore also resampled onto a common progress coordinate and represented by the normalized current profile

\begin{equation}
i(\tau)=\frac{I(\tau)}{\max |I|}
\label{eq:norm_current}
\end{equation}

where $\tau$ is the normalized progress through the selected cycle, operating window, or source-data trace. The 2025 EV fleet trace is placed only in Segment~1 of Fig.~\ref{fig:patterns} because the accessible source contains a single charging trace rather than repeated operating sections. To quantify operational variability and dynamic loading characteristics, the duty-structure index (DSI) is defined as

\begin{equation}
\begin{aligned}
\mathrm{DSI} &= \sqrt{A^2+R^2} \\
A &= \mathrm{std}(i) \\
R &=
\ln\!\left(
1+\frac{1}{2M}
\sum_{m=1}^{M}\sum_j
|i_m(\tau_{j+1})-i_m(\tau_j)|
\right)
\end{aligned}
\label{eq:dsi}
\end{equation}

where $M$ denotes the number of selected records. Here, $A$ characterizes the dispersion of the normalized current profile, whereas $R$ captures cumulative ramping behavior with logarithmic compression to reduce sensitivity to extreme transients. 

\section{Results}

\subsection{Duty Structure and Cycle Response}

\begin{figure}[!t]
\centering
\includegraphics[width=0.90\columnwidth]{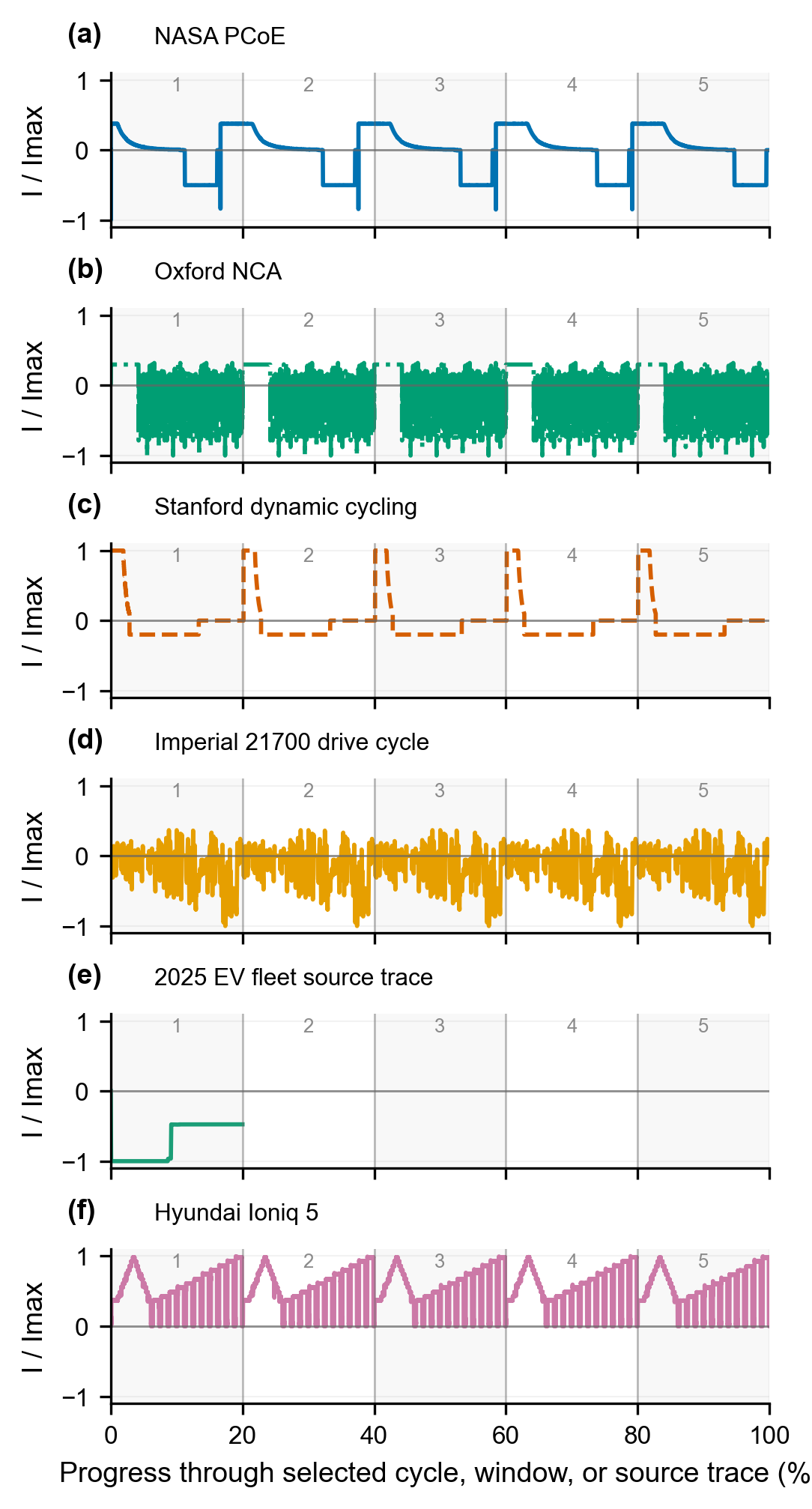}
\caption{Normalized current profiles over selected cycles, operating windows, or source-data traces. Note: The 2025 EV fleet trace occupies only segment 1 because repeated sections are not available.}
\label{fig:patterns}
\end{figure}

Fig.~\ref{fig:patterns} compares the behavior of the battery testing profile and the field applications, presenting the difference between the cycling tests in \textbf{a} and \textbf{c} and the dynamic operation tests in \textbf{b} and \textbf{d} of the subfigures.
The 2025 EV fleet panel places its single accessible pack-level charging trace in segment~1 and leaves segments~2--5 empty; the Hyundai panel uses five repeated charger-control blocks.
The contrast is immediate. The NASA profile now shows repeated charge-discharge laboratory cycles, with long controlled charging stages and nearly constant-current discharges. The Oxford window repeats the available application-shaped current profile and introduces frequent current variation during the discharge stage. The Stanford window reveals repeated long phases of charge, discharge, and rest. The Imperial 21700 trace uses five 1800~s repeat windows from the WLTP-based setting. The 2025 EV fleet source trace shows a pack-level transition from a high-current stage to a lower-current stage, while the Hyundai charging record shows repeated charger-control blocks.
\begin{figure}[!t]
\centering
\includegraphics[width=0.92\columnwidth]{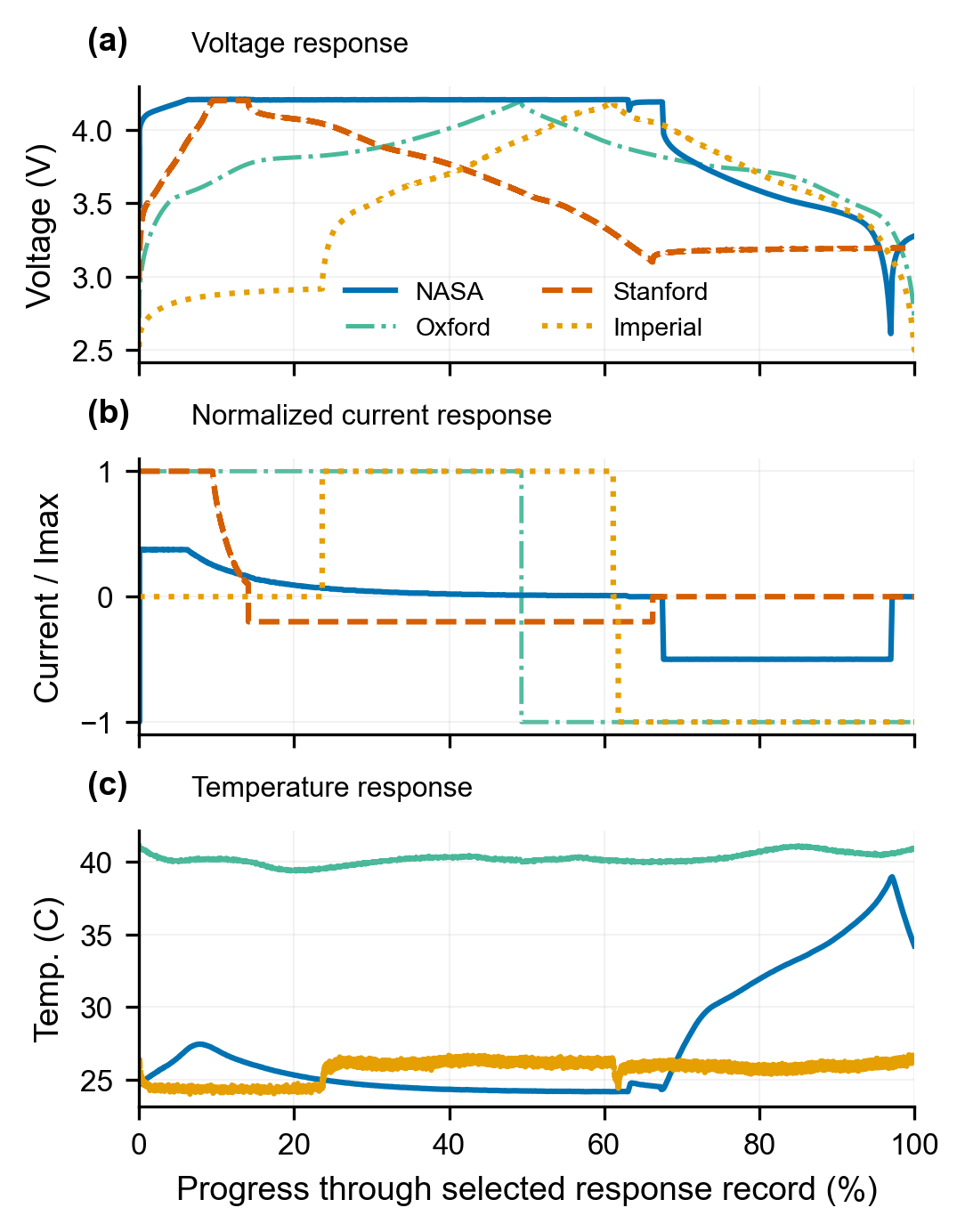}
\caption{Voltage, normalized-current, and temperature responses for selected cell-level records in reference performance tests (RPTs)}
\label{fig:cyclecompare}
\end{figure}

Looking into the diagnosis cycle, where stable charging-discharging processes are implemented, Fig.~\ref{fig:cyclecompare} extends the comparison from the duty structure to the measured response by using normalized progress through each selected response record. Reference performance tests (RPTs) are periodic diagnostic tests used during ageing campaigns to measure capacity retention under controlled conditions. NASA is shown with a charge--discharge pair, Oxford with the application-shaped example cycle, Stanford with the selected dynamic cycle, and Imperial with the available RPT0 diagnostic response because the public drive-cycle setting does not include a matching voltage--temperature trace. Voltage is retained in physical units, current is normalized by peak magnitude, and temperature is shown if measured.

\subsection{Short-Term and Long-Term insights}

\begin{figure}[!t]
\centering
\includegraphics[width=0.92\columnwidth]{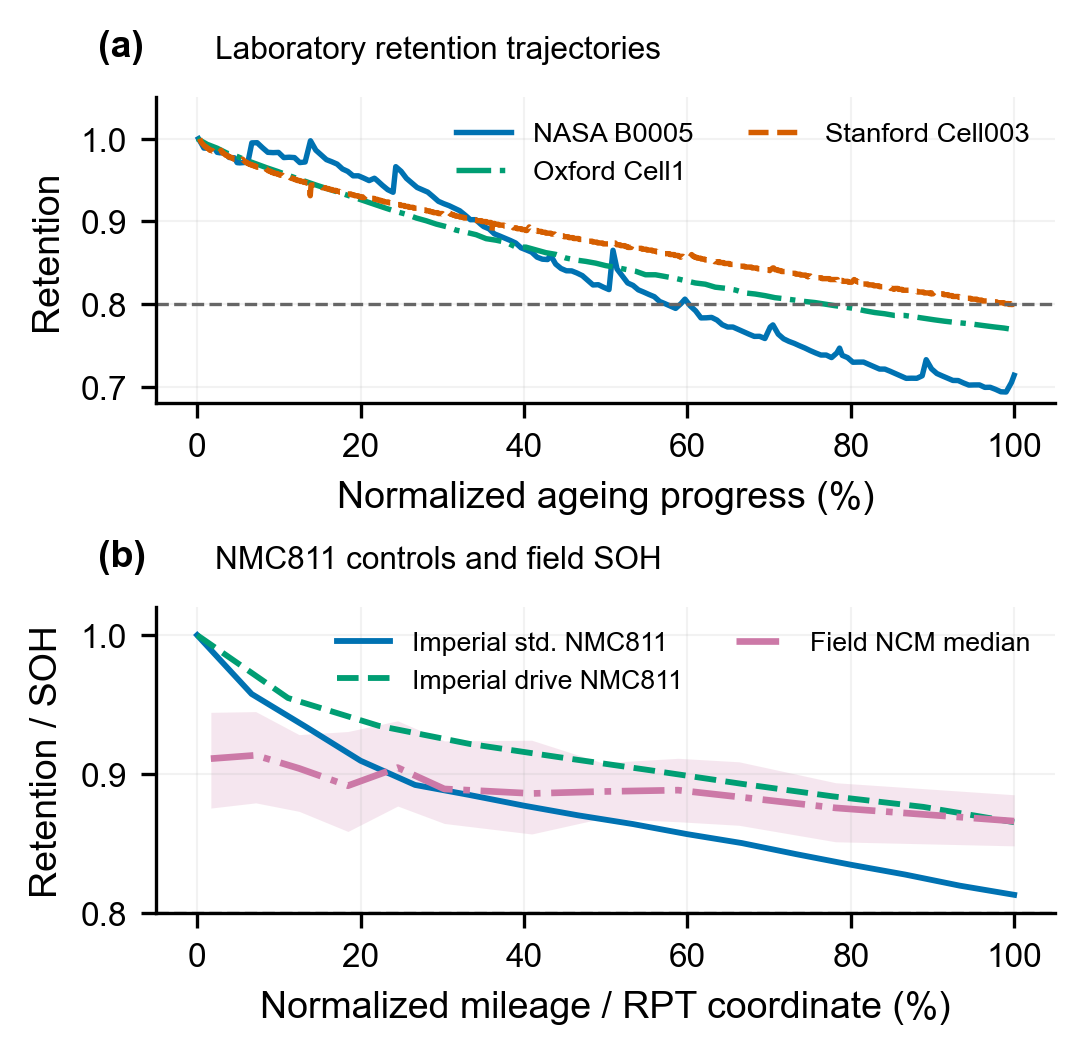}
\caption{Long-term retention and SOH comparison across laboratory datasets and chemistry-aligned evidence using normalized aging progress.}
\label{fig:degradation}
\end{figure}

Fig.~\ref{fig:degradation}(a) compares long-term degradation trajectories from NASA, Oxford, and Stanford using normalized retention on the vertical axis and normalized ageing progress on the horizontal axis. The original ageing coordinates are cycle count for NASA and Stanford and checkpoint count for Oxford, but they are mapped to 0--100\% to compare trajectory shape. NASA shows the faster decline with more fluctuation in the selected records, Oxford degrades more gradually, and Stanford remains the shallowest of the three until late in life.

Fig.~\ref{fig:degradation}(b) adds a chemistry-aligned comparison between laboratory NMC811 controls and real-field NCM outcomes. The Imperial 21700 controls at $25\,^{\circ}\mathrm{C}$~\cite{imperial} show that the drive-cycle laboratory condition retains 0.865 by RPT9, whereas the standard-cycle control falls to 0.813 by RPT15. These RPT coordinates and the field mileage coordinate are also normalized to 0--100\% in the figure. The field NCM source-data release~\cite{fieldsource} reports a median SOH trajectory in a similar numerical band to the milder drive-cycle laboratory condition, but its interquartile spread reflects fleet heterogeneity that laboratory single-cell controls do not capture, but is evident.

\begin{figure}[!t]
\centering
\includegraphics[width=0.90\columnwidth]{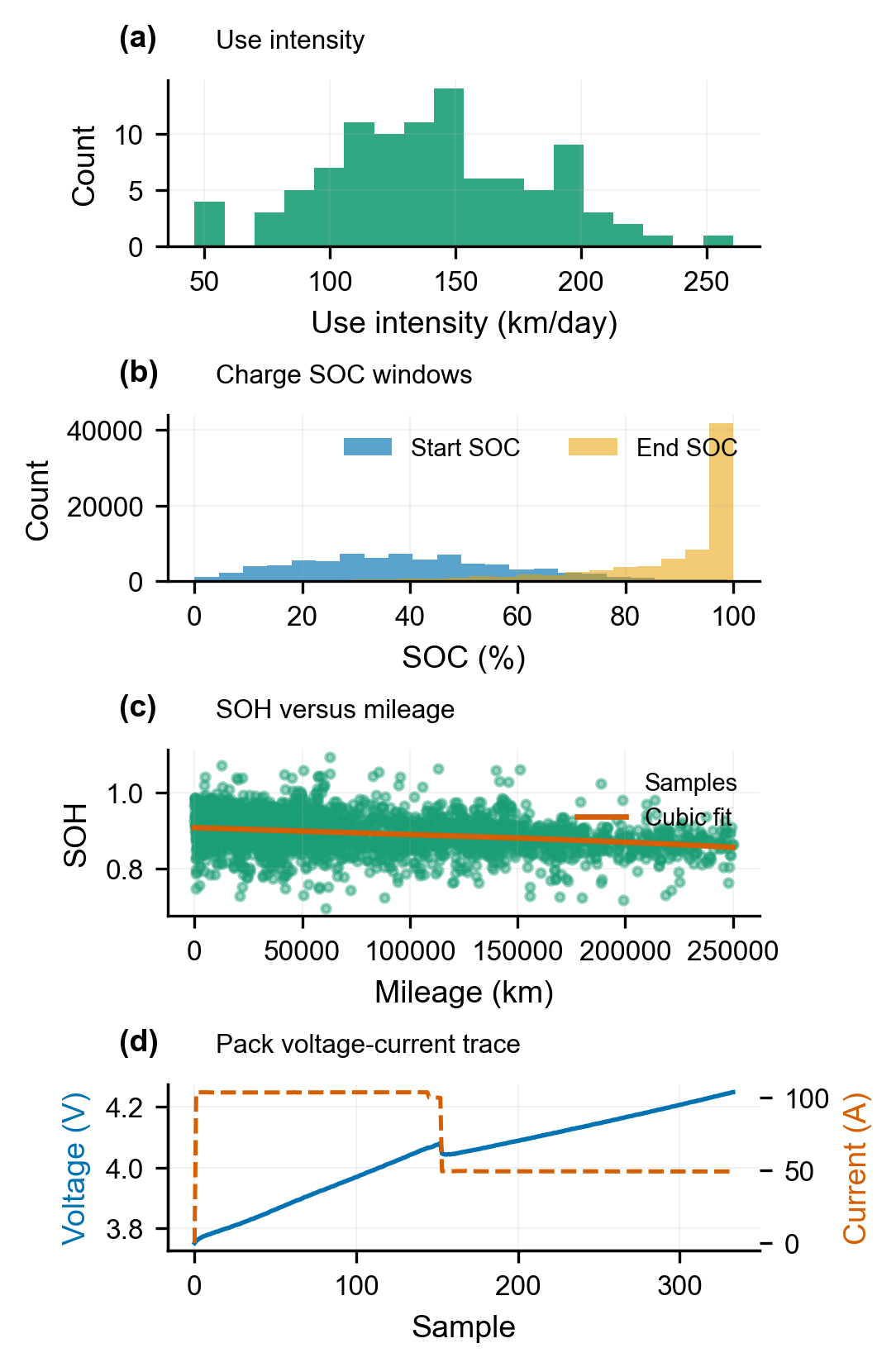}
\caption{Field-EV evidence to characterize real operating diversity.}
\label{fig:field}
\end{figure}

Fig.~\ref{fig:field} complements the single duty trace with broader field context from Liu \emph{et al.}~\cite{fieldsource}. The source-data release shows a median use intensity of 137.2~km/day, with a 10th--90th percentile range of 90.6 to 197.6~km/day. Charge-window statistics show a median charge start SOC of 38.0\% and a median charge end SOC of 96.0\%, with 56.9\% of charging events ending at or above 95\% SOC. The SOH-versus-mileage panel shows an overall downward trend but also substantial spread, indicating that mileage alone does not explain field health outcomes. The selected Hyundai Ioniq~5 charging session from the KIT record~\cite{kitdataset} lasts 11.30~h, reaches a peak summed phase current of 48.1~A, and increases SOC by 53.0 percentage points.

\section{Discussion}

The results show that battery test patterns are not merely different experimental formats; they are different kinds of evidence. Fig.~\ref{fig:patterns} and Fig.~\ref{fig:cyclecompare} show that the controlled NASA cycle, the Oxford drive-cycle discharge, the Stanford dynamic protocol, the Imperial NMC811 drive-cycle setting, the 2025 EV fleet source trace, and the Hyundai charging trace differ in temporal structure even after normalization. For the repeated records, the quantitative indices in Table~\ref{tab:deg_drift} are calculated from one representative segment: one NASA charge--discharge pair, one Oxford example cycle, one Stanford dynamic cycle, one Imperial 1800~s drive-cycle window, and one Hyundai charger-control block. The corresponding DSI is 0.699 for NASA, 0.639 for Stanford, 2.936 for Oxford, 2.855 for Imperial, 0.630 for the single 2025 EV fleet source trace, and 2.268 for the selected Hyundai charger-control block. The usage metrics add a complementary interpretation: Oxford has the highest usage intensity and usage C-rate among the selected cell records, whereas Stanford is active for a substantial fraction of time but at a low average C-rate. Thus, the most structured profiles are not necessarily those with the largest amplitude spread; they are the profiles with repeated ramps, control actions, and transient current changes.

The long-term comparison leads to the same conclusion at the ageing level. In Fig.~\ref{fig:degradation}(a), the selected laboratory records reach the 80\% region on very different interpolated horizons: near cycle 351 for NASA, near checkpoint 6292 for Oxford, and near cycle 1019 for Stanford. These are not just different lifetimes; they are different ageing trajectories shaped by different test logic. Fig.~\ref{fig:degradation}(b) extends this argument by showing that chemistry-aware alignment improves interpretation but does not eliminate the field gap. The Imperial NMC811 control retains 0.813 under standard cycling and 0.865 under drive-cycle ageing, whereas the field NCM source reports a median SOH of 0.889 together with visible dispersion. The implication is that even a chemistry-matched laboratory reference cannot stand in for the diversity of field operation, because the field record retains the effects of heterogeneous charge windows, operating intensity, and control history.

The field statistics make that diversity explicit. The accessible EV source data show a median use intensity of 137.2~km/day, a median charging window from 38.0\% to 96.0\% SOC, and 56.9\% of charging events ending at or above 95\% SOC. Those operating patterns are poorly represented by one canonical laboratory cycle. Table~\ref{tab:deg_drift} condenses the main quantitative implications: the NASA case provides the controlled baseline and fastest relative fade among the selected cell datasets, the Oxford and Imperial records show the largest single-segment duty-structure indices among the laboratory traces, the fleet source combines a low-DSI single trace with broad operating and SOH dispersion, and the Hyundai session shows repeated charger-control structure. For application-oriented studies, the relevant question is which application claim a given test can defend. Controlled cycling is appropriate for mechanism isolation and stable benchmarking. Drive-cycle and dynamic protocols are better suited to claims about application-shaped response. Field traces and field health statistics are required when the claim concerns operational realism, fleet heterogeneity, or transfer to deployed systems.
\begin{table}[!t]
\caption{Key quantified duty-structure, degradation, and operating-drift indicators across representative single segments and source records.}
\label{tab:deg_drift}
\centering
\scriptsize
\setlength{\tabcolsep}{3pt}
\renewcommand{\arraystretch}{1.12}

\begin{tabular*}{\columnwidth}{@{\extracolsep{\fill}}
p{1.05cm}
c
c
c
c
p{1.5cm}
p{3.0cm}
@{}}
\toprule
Source & DSI & UF & UI & UC & Degradation & Main implication \\
\midrule

NASA
& 0.699
& 0.70
& 0.296
& 0.29
& 80\%@351 cycles
& Fastest degradation under controlled high-intensity cycling. \\

Oxford
& 2.936
& 0.75
& 0.817
& 2.00
& 80\%@6292 checkpoints
& Highest sustained usage C-rate with longest cycling lifetime. \\

Stanford
& 0.639
& 0.66
& 0.082
& 0.40
& 80\%@1019 cycles
& Smooth low-C-rate cycling with reduced operational stress. \\

Imperial
& 2.855
& 0.77
& 0.150
& 0.14
& $\Delta$ret. +0.052 at RPT
& Highest ramp-dominated duty structure and transient variability. \\

Hyundai
& 2.268
& 0.84
& 0.132
& 0.26
& n.a.
& No degradation label; full-session $\Delta$SOC 53.0. \\

EV fleet
& 0.630
& 1.00
& --
& 0.48
& SOH$_{50}$ 0.889; cycles n.r.
& Median field SOH; cycle count not reported. \\

\midrule
\multicolumn{7}{@{}p{\columnwidth}@{}}{
\textit{Note:} DSI, UF, UI, and UC are calculated from one representative segment for repeated records; degradation indicators use the full ageing or fleet-health source.
UC is calculated over active charging time following the prior usage C-rate definition.
Hyundai usage C-rate is estimated using an AC-current proxy.
EV-fleet UC is sample-coordinate based and UI is not reported because timestamps and cycle counts are unavailable.
} \\
\bottomrule
\end{tabular*}
\end{table}
\section{Conclusion}

This paper quantified the laboratory-to-field duty-profile gap using controlled, drive-cycle, dynamic, chemistry-aligned, and field EV evidence. The selected records are not interchangeable benchmarks: they support different claims because they impose different current structures, diagnostic conditions, and operating histories.

The duty-profile comparison gives the clearest numerical result. The single-segment DSI is 0.699 for NASA, 0.639 for Stanford, 2.936 for Oxford, 2.855 for Imperial, 0.630 for the EV-fleet source trace, and 2.268 for the Hyundai charger-control block. The usage metrics show a similar separation: Oxford has the highest usage intensity and usage C-rate among the selected cell records (UI = 0.817, UC = 2.00), while Stanford has lower active cycling intensity (UI = 0.082, UC = 0.40). These values confirm that laboratory protocols differ not only in scale, but also in temporal structure.

The degradation comparison shows that the interpretation of ageing is also duty-pattern dependent. The 80\% retention region appears near 351 NASA cycles, 6292 Oxford checkpoints, and 1019 Stanford cycles. In the NMC/NCM comparison, Imperial retains 0.813 under standard cycling and 0.865 under drive-cycle ageing, whereas the field source reports median SOH of 0.889 with an interquartile spread. The field data also show operational diversity: median use intensity is 137.2~km/day, the median charging window is 38.0--96.0\% SOC, and 56.9\% of charges end at or above 95\% SOC.

The practical conclusion is that laboratory testing is necessary but not sufficient for application-oriented claims. Battery studies should report chemistry, capacity, temperature, and cycle count together with charge/discharge windows, temporal variability, rest structure, usage frequency, usage intensity, usage C-rate, and the intended application analogy. Without this alignment, a performance or degradation metric can be valid for its test protocol while remaining weak evidence for deployed operation.

\balance


\begin{thebibliography}{99}

\bibitem{zhao2023bess}
C. Zhao, P. B. Andersen, C. Tr{\ae}holt, and S. Hashemi, ``Grid-connected battery energy storage system: A review on application and integration,'' \textit{Renewable and Sustainable Energy Reviews}, vol. 182, p. 113400, Aug. 2023.

\bibitem{geslin2025dynamic}
A. Geslin, L. Xu, D. Ganapathi \textit{et al.}, ``Dynamic cycling enhances battery lifetime,'' \textit{Nature Energy}, vol. 10, pp. 172--180, 2025, doi: 10.1038/s41560-024-01675-8.

\bibitem{nasa}
B. Saha and K. Goebel, ``Battery Data Set,'' NASA Prognostics Data Repository, NASA Ames Research Center, Moffett Field, CA, USA, 2007.

\bibitem{oxforddataset}
D. Howey and C. Birkl, ``Oxford Battery Degradation Dataset 1,'' University of Oxford, Oxford, U.K., 2017, doi: 10.5287/bodleian:KO2kdmYGg.

\bibitem{birkl}
C. R. Birkl, M. R. Roberts, E. McTurk, P. G. Bruce, and D. A. Howey, ``Degradation diagnostics for lithium ion cells,'' \emph{Journal of Power Sources}, vol. 341, pp. 373--386, 2017, doi: 10.1016/j.jpowsour.2016.12.011.

\bibitem{severson}
K. A. Severson \emph{et al.}, ``Data-driven prediction of battery cycle life before capacity degradation,'' \emph{Nature Energy}, vol. 4, no. 5, pp. 383--391, 2019, doi: 10.1038/s41560-019-0356-8.

\bibitem{stanford}
A. Geslin, L. Xu, D. Ganapathi, K. Moy, W. C. Chueh, and S. Onori, ``Dynamic cycling enhances battery lifetime,'' \emph{Nature Energy}, vol. 10, pp. 172--180, 2025, doi: 10.1038/s41560-024-01675-8.

\bibitem{fieldsource}
H. Liu, C. Li, X. Hu, J. Li, K. Zhang, Y. Xie, R. Wu, and Z. Song, ``Multi-modal framework for battery state of health evaluation using open-source electric vehicle data,'' \emph{Nature Communications}, vol. 16, Art. no. 1137, 2025, doi: 10.1038/s41467-025-56485-7.

\bibitem{imperial}
N. Kirkaldy, M. A. Samieian, G. J. Offer, M. Marinescu, and Y. Patel, ``Lithium-ion battery degradation: Comprehensive cycle ageing data and analysis for commercial 21700 cells,'' \emph{Journal of Power Sources}, vol. 603, p. 234185, 2024, doi: 10.1016/j.jpowsour.2024.234185.

\bibitem{kitdataset}
S. Beichter \emph{et al.}, ``Open-source data set for characterizing the charging behavior of electric vehicles,'' Zenodo, Feb. 9, 2026, doi: 10.5281/zenodo.17866847.


\end{thebibliography}
\end{document}